\documentclass[preview,3p]{elsarticle}
\usepackage[title]{appendix}
\usepackage{lineno,hyperref,natbib}
\usepackage{amsmath,amssymb}
\usepackage{float}
\usepackage{graphicx}
\usepackage{subfigure}
\usepackage{mathtools}
\usepackage{algpseudocode} 
\usepackage{CJK}  
\usepackage{booktabs}  
\usepackage{threeparttable}  
\usepackage{breakurl}
\usepackage{multirow}
\usepackage[linesnumbered,boxed,ruled]{algorithm2e}
\SetKwProg{Fn}{Subalgorithm}{}{}
\newcommand{\rmnum}[1]{\romannumeral #1}
\newcommand{\Rmnum}[1]{\expandafter\@slowromancap\romannumeral #1@}
\modulolinenumbers[5]
\journal{Journal of \LaTeX\ Templates}

\begin{document}
\begin{frontmatter}
\title{VDPC: Variational Density Peak Clustering Algorithm}
\author[mymainaddress,my3address]{Yizhang Wang}
\author[my4address]{Chai Quek}
\author[my5address]{You Zhou}
\author[my1address,my2address]{Di Wang\corref{mycorrespondingauthor}}

\ead{wangdi@ntu.edu.sg}
\cortext[mycorrespondingauthor]{Corresponding author}
\address[mymainaddress]{College of Information Engineering, Yangzhou 
University, Yangzhou, China}
\address[my3address]{Institute of Scientific and Technical Information of China, Beijing, China}
\address[my4address]{School of Computer Science and Engineering, Nanyang Technological University, Singapore}
\address[my5address]{College of Computer Science and Technology, Jilin 
	University, Changchun, China}
\address[my1address]{Joint NTU-UBC Research Centre of Excellence in Active Living for the Elderly, Nanyang Technological University, Singapore}
\address[my2address]{Joint NTU-WeBank Research Centre on Fintech, Nanyang Technological University, Singapore}

\begin{abstract}
The widely applied density peak clustering (DPC) algorithm makes an intuitive cluster formation assumption that cluster centers are often surrounded by data points with lower local density and far away from other data points with higher local density. However, this assumption suffers from one limitation that it is often problematic when identifying clusters with lower density because they might be easily merged into other clusters with higher density.
As a result, DPC may not be able to identify clusters with variational density. 
To address this issue, we propose a variational density peak clustering (VDPC) algorithm, which is designed to systematically and autonomously perform the clustering task on datasets with various types of density distributions.
Specifically, we first propose a novel method to identify the representatives among all data points and construct initial clusters based on the identified representatives for further analysis of the clusters' property. Furthermore, we divide all data points into different levels according to their local density and propose a unified clustering framework by combining the advantages of both DPC and DBSCAN. Thus, all the identified initial clusters spreading across different density levels are systematically processed to form the final clusters.
To evaluate the effectiveness of the proposed VDPC algorithm, we conduct extensive experiments using 20 datasets including eight synthetic, six real-world datasets and six image datasets. The experimental results show that VDPC outperforms two classical algorithms (i.e., DPC and DBSCAN) and four state-of-the-art extended DPC algorithms.

\end{abstract}


\end{frontmatter}


\section{Introduction}
\begin{figure}[!t]
	\centering
	\includegraphics[scale=0.5]{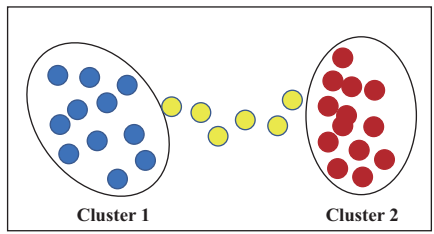}
	\caption{An illustration of having connected points (yellow points) between two dense natural clusters.}
	\label{figconnect}
\end{figure}

Clustering is an important unsupervised knowledge acquisition method, which divides the unlabeled data into different groups \cite{atilgan2021efficient,d2021automatic}. Different clustering algorithms make different assumptions on the cluster formation, thus, most clustering algorithms are able to well handle at least one particular type of data distribution but may not well handle the other types of distributions. For example, K-means identifies convex clusters well \cite{bai2017fast}, and DBSCAN is able to find clusters with similar densities \cite{DBSCAN}. 
Therefore, most clustering methods may not work well on data distribution patterns that are different from the assumptions being made and on a mixture of different distribution patterns. Taking DBSCAN as an example, it is sensitive to the loosely connected points between dense natural clusters as illustrated in Figure~\ref{figconnect}. The density of the connected points shown in Figure~\ref{figconnect} is different from the natural clusters on both ends, however, DBSCAN with fixed global parameter values may wrongly assign these connected points and consider all the data points in Figure~\ref{figconnect} as one big cluster.

Density peak clustering (DPC) is a recently proposed clustering algorithm \cite{rodriguez2014clustering}, which receives more and more attention \cite{wang2020mcdpc,hou2020density,abbas2021denmune}. DPC makes a relatively novel assumption on the cluster formation that cluster centers are often surrounded by data points with lower density and they are also far away from other data points with higher density \cite{rodriguez2014clustering}. 
This cluster formation assumption leads to a series of advantages of DPC, namely connected points between different natural clusters are easily identified (assigning the connected points to the nearest density peak clusters, respectively), clusters are determined in a non-iterative manner, outliers are easily identified (naturally by DPC's assumptions), etc.

\begin{figure}[!t]
	\centering
	\subfigure[  ]{
		\label{comgoround}
		\includegraphics[width=0.4\textwidth]{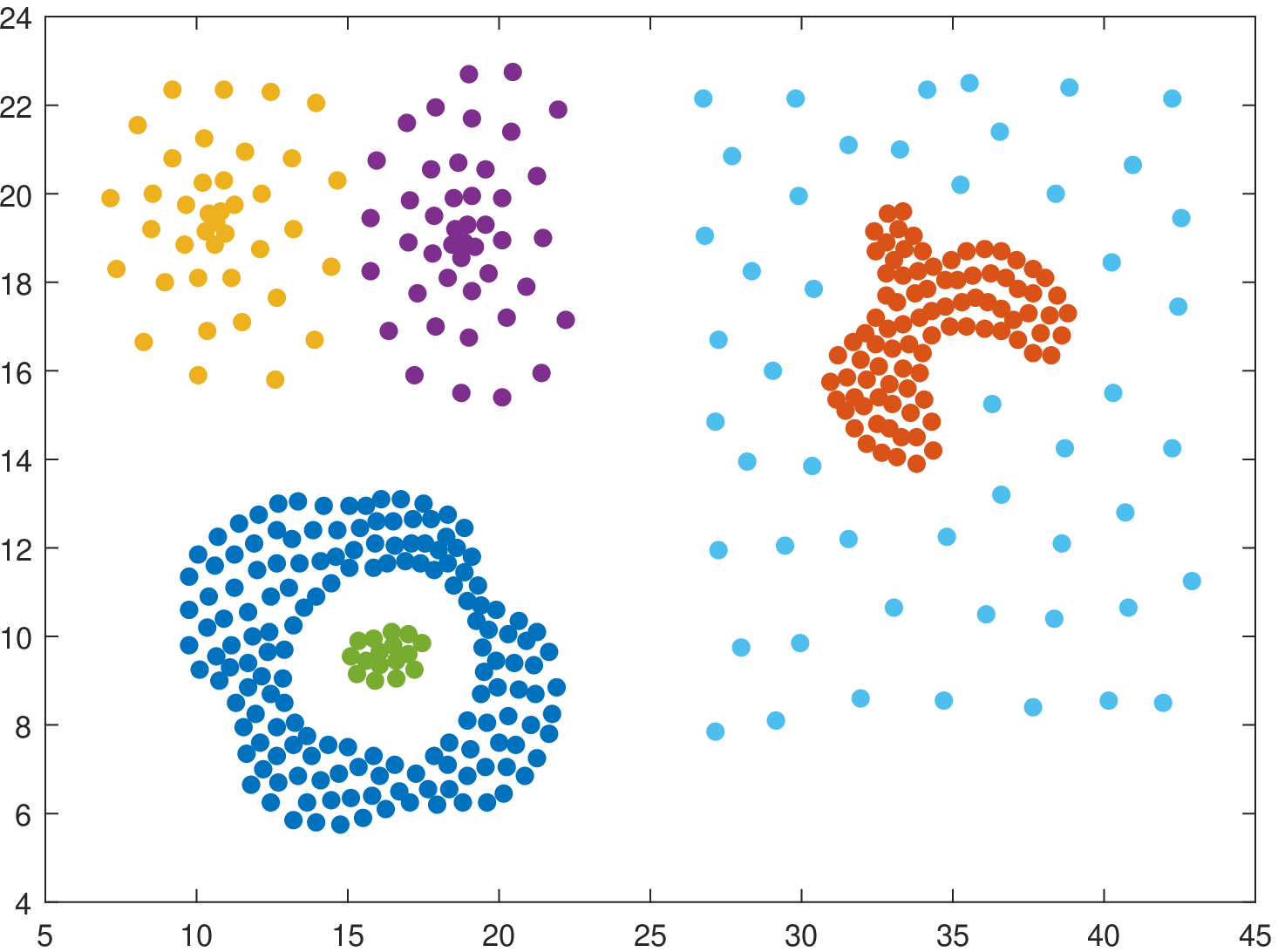}}
	\subfigure[]{
		\label{Fig2b}
		\includegraphics[width=0.4\textwidth]{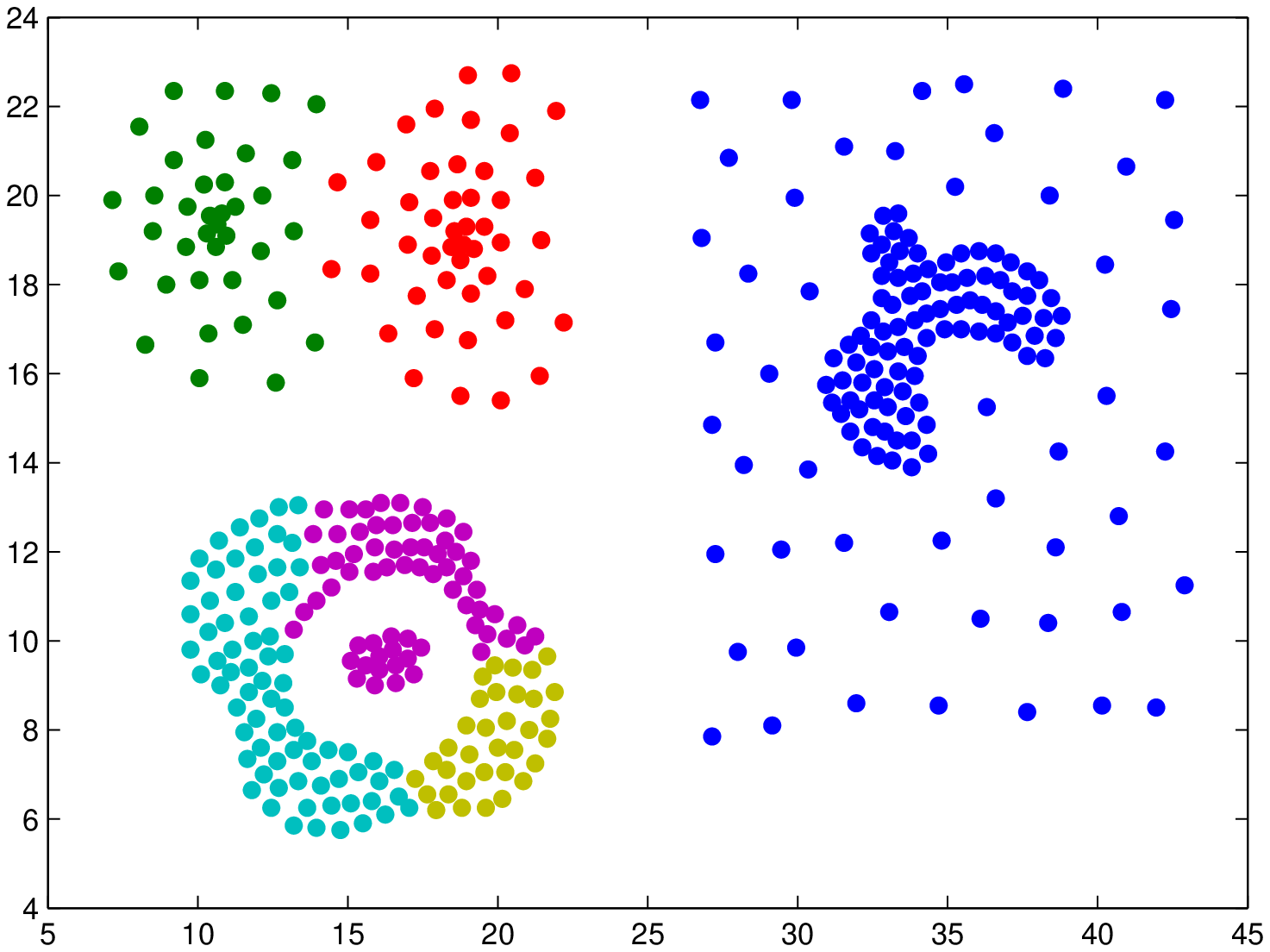}}	
	\caption{Visualization of the Compound dataset and the clustering result generated by DPC. (a) The ground-truth of a typical variational density dataset Compound \cite{Chang2008Robust}. (b) The clustering result of DPC on dataset Compound, the two natural clusters with different density are merged into one cluster (the dark blue cluster on the right) by DPC.}
	\label{Fig1}
\end{figure}

However, DPC suffers from one major limitation: 
DPC may not identify clusters with relatively lower density. As illustrated in Figure~\ref{Fig1}, for the  Compound  dataset \cite{Chang2008Robust}, which comprises natural clusters  with variational density, DPC obtains inferior results. The light blue (low density) and red (high density) natural clusters on the right-hand side of Figure~\ref{comgoround} are indistinguishable by DPC (see Figure~\ref{Fig2b}).

The afore-illustrated drawback of DPC makes it difficult to identify clusters with variational density. 
To better solve this issue and strive for better performance, in this paper, we propose a variational density peak clustering (VDPC) algorithm. Our proposed VDPC algorithm takes the advantage of both DPC and DBSCAN. Specifically,
if each natural cluster in the underlying dataset has similar density, it is efficient and straightforward to identify all these clusters by applying DBSCAN \cite{DBSCAN} thanks to its cluster formation assumption (see Section \ref{dbscan}).
By extending the capability of DBSCAN, for a variational density dataset, we first determine the different density levels of the identified initial cluster centers based on DPC's local density analysis (see Figure~\ref{lowhigh}). We then identify the ultimate cluster centers in a divide-and-conquer approach in their respective density levels.
To systematically and autonomously determine the different density levels of the identified initial cluster centers, we propose a novel method to let VDPC self-determine the number of density levels exist in the underlying dataset.   
Furthermore, we conduct extensive experiments on both synthetic and real-world datasets with single or multiple density levels. The experimental results show that our proposed VDPC algorithm outperforms DPC, DBSCAN and the state-of-the-art DPC variants.
Our contributions in this paper are as follows:
\begin{figure}[!t]
	\centering
	\includegraphics[scale=0.3]{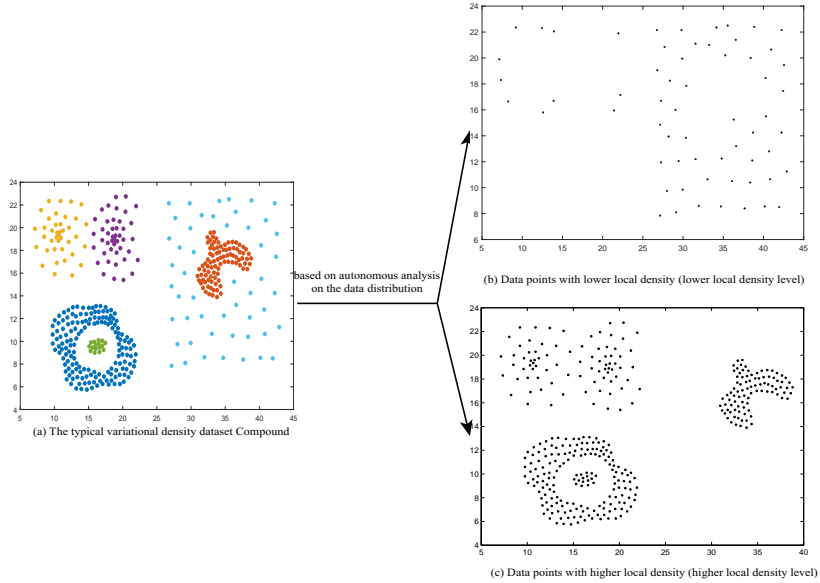}
	\caption{Illustration on how VDPC differentiates different density levels (see more details in Section \ref{vdpc}).}
	\label{lowhigh}
\end{figure}

(\rmnum{1}) We propose a novel data distribution analysis method, which can systematically divide all the data points in a given dataset into respective density levels. Based on such autonomously obtained density levels, we can get more insights on the data distribution patterns of the dataset to further identify the proper subsequent cluster formation strategy.

(\rmnum{2}) Our proposed VDPC algorithm requires a small number of predetermined parameters, i.e., two, the same number as required by DBSCAN and one less than DPC. Moreover, the complexity of VDPC is on the same magnitude of DPC and DBSCAN.

(\rmnum{3}) We propose a novel dynamic cluster formation strategy, which combines the advantages of both DPC and DBSCAN. As such, our cluster formation strategy adapts according to the analyzed data distribution patterns to minimize the possibility of wrongly assigning the lower density points.

(\rmnum{4}) By evaluating the effectiveness of VDPC using both widely adopted synthetic and real-world datasets with performance comparisons, we show that VDPC outperforms other clustering methods, especially on datasets comprising multiple density levels.

The remainder of this paper is organized as follows. In Section \ref{rw}, we review the density peak clustering algorithm, its extensions, and other clustering methods. In Section \ref{vdpc}, we provide the technical details of VDPC with examples and illustrations. In Section \ref{erad}, we present the experimental results with discussions. In Section \ref{cafw}, we conclude this paper and propose future work.

\section{Related Work}
\label{rw}
In this literature review section, we introduce the fundamentals of DPC, recent extensions of DPC, and other clustering methods. In addition, we discuss the pros and cons of these reviewed methods and present how we propose VDPC by combining the advantages of prior models.

\subsection{Density Peak Clustering (DPC) Algorithm}
\label{dpc}
DPC articulates that each data point has two properties, namely $\rho$ as the local density of individual data points and  $\delta$ as the minimum distance between one data point and another with higher density. For $N$ data points in a dataset $D=(x_1, x_2,\cdots,x_N )^T$, the similarity between data points $x_i$ and $x_j$ is defined as  
\begin{equation} \label{educa}
S(x_i,x_j)=\left \lVert x_i-x_j \right\rVert,
\end{equation}
where $|| \cdot ||$ denotes the Euclidean distance.
The upper triangular similarity matrix of $S$, denoted as $U$, is defined as 
\begin{equation}\label{upper}
	U=\left(\begin{array}{cccc}
	S_{11} & S_{12} & \cdots & S_{1N} \\
	&S_{22}& \cdots &S_{2N}\\
	&& \ddots & \vdots \\
	&&&S_{NN}
	\end{array}\right),
\end{equation}
where the elements in $U$ is arranged in the ascending order to form a one-dimensional vector $u$=$(u_1,u_2$,$\cdots$, $u_{N(N-1)/2})$. Then, the local density ($\rho$) of data point $x_i$ is defined as 
\begin{equation} \label{rho}
\rho_i=\sum_{j}e^{-(\frac{\left \lVert x_i-x_j \right \rVert}{d_c})^2},
\end{equation}
\begin{equation} \label{dc}
d_c=u_{\textit{round}(pct\%*N(N-1)/2)},
\end{equation}
where $\textit{round}$ $(\cdot)$ denotes to round a certain decimal up to the nearest integer. 
Therefore, the value of $\rho$ is a dependent of $pct$, which is a parameter needs to be predefined in DPC. In essence, \textit{pct} is the cut-off value to define the neighborhood radius used for computation of local density.

Subsequently, parameter $\delta$ of data points $x_i$, which denotes the minimum distance between one data point and another with higher density, is defined as
\begin{equation} \label{delta}
\delta_i= \min_{j:\rho_j>\rho_i} S(x_i,x_j).
\end{equation}

\begin{figure}[!t]
	
	\centering
	\includegraphics[scale=0.5]{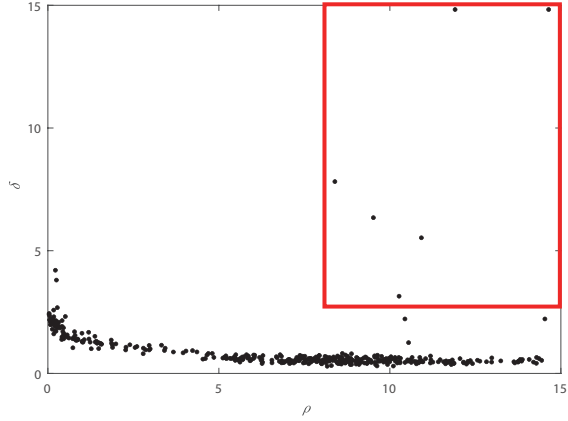}
	\caption{The decision graph of dataset Compound generated by DPC (\textit{$pct=1.9$}), the data points in the red rectangle are selected as cluster centers.}
	\label{figoutliers}
\end{figure}

\noindent It is obvious that once the similarity matrix of $S$ is determined (see (\ref{upper})), the values of $\delta_i$ are subsequently determined. 
The values of $\rho_i$ and $\delta_i$ of all the data points are then used to plot the 2D  decision graph (see Figure~\ref{figoutliers}). 
According to DPC's assumption, cluster centers (i.e., density peaks) should have both high $\rho$ and high $\delta$ values. Based on the decision graph, users are required to manually identify cluster centers by specifying a rectangle on the decision graph (see the red rectangle in Figure~\ref{figoutliers}), whereby the data points inside this red rectangle are the user-identified cluster centers.
To specify the rectangle of cluster centers, users need to provide the coordinates of the bottom-left vertex, denoted as ($\rho_u, \delta_u$).  Therefore, DPC requires users to set the values of three parameters, namely \textit{pct}, $\rho_u$ and $\delta_u$.

Once the cluster centers are manually identified, each remaining data point is sequentially assigned to its nearest higher density neighbor so as to form all clusters. 
For example, the clustering results shown in Figure~\ref{Fig2b} are obtained by the identified cluster centers shown in Figure~\ref{figoutliers}.

By comparison with DPC, our proposed VDPC automates the process of identifying cluster centers and further extends the capability of handling clusters with variational density.

\subsection{Shared Nearest Neighbour Clustering Algorithm (SNNC)}
\label{mknn}

$k$ Nearest Neighbour (KNN) is a well-established classification and regression method. Shared Nearest Neighbour clustering algorithm (SNNC) is built upon KNN, which only requires one user-defined parameter $k$ (number of nearest neighbours) \cite{patidar2012analysis}. For two data points $x_i$ and $x_j$, $knn(x_i)$ denotes $k$ nearest neighbours of $x_i$ and $knn(x_j)$ denotes $k$ nearest neighbours of $x_j$, where the neighbours are identified based on the Euclidean distance. If $knn(x_i)$ and $knn(x_j)$ have more than one shared nearest neighbours, data points $x_i$ and $x_j$ are deemed as belonging to one cluster. 
The overall cluster formation process of SNNC is shown in Algorithm~\ref{algmknn}. SNNC has been widely adopted to improve the performance of other clustering algorithms (e.g., \cite{mehta2017segmentation,ros2019munec}).

\begin{algorithm}[htbp]
	\KwIn{ dataset $D$, user-defined parameter $k$, and adjacency matrix $A$.}
	\KwOut{clustering results}
	$A(i,j)=\begin{cases}
	1, & knn(x_i) \cap knn(x_j) >1\\ 
	0, & otherwise
	\end{cases}$\\ 
	$N$$ \leftarrow $ the number of data points in \textit{D}\;
	\For{$i=1$$ \rightarrow $ $N$ }   
	{
		\For{$j=1$$ \rightarrow $$N$ }   
		{
			\eIf{$A(i,j)=1$}{
				data points  $x_i$ and $x_j$ belong to the same cluster\;
				
			}{
				data points  $x_i$ and $x_j$ belong to different clusters\;
			}			
		}		
	}
    each remaining data point that does not belong to any cluster becomes a cluster\; 
	output the clustering results.
	\caption{The shared nearest neighbour clustering algorithm (SNNC)}
	\label{algmknn}
\end{algorithm}

In this paper, we  propose an extension of SNNC (see Section \ref{num2}) to enable VDPC to efficiently determine the cluster formation strategy in certain circumstances.

\subsection{DBSCAN}
\label{dbscan}
DBSCAN is a classical density-based clustering algorithm, which requires two important parameters: \textit{Eps} as the radius of underlying neighborhood and \textit{MinPts} as the minimum number of data points within the neighborhood \cite{DBSCAN}. 
DBSCAN has the following definitions: Data point $x_p$ is a core point  if at least \textit{MinPts} data points have the distance of less than \textit{Eps} away from $x_p$ (including data point $x_p$ itself); A data point $x_p$ is directly density-reachable from a core point $x_q$ if $x_p$ is within distance \textit{Eps} from $x_q$; A data point $x_p$ is density-reachable from another data point $x_q$ if there is a chain of data points
$\{x_p,\cdots,x_o,x_{o+1},\cdots,x_q\}$ such that any data point $x_{o+1}$ is directly density-reachable from $x_o$ in this chain.
Finally, DBSCAN generates clusters as follows: If $x_p$ is a core point, then it forms a cluster together with all data points that are density-reachable from it.

The performance of DBSCAN heavily depends on the predetermined parameter values. Specifically,
if a larger value of \textit{Eps} is used, the number of noises (outliers that do not belong to any dense cluster) is generally smaller; if a larger value of \textit{MinPts}  is used, the number of noises is generally larger.
Through conducting extensive experiments, we find that DBSCAN is able to accurately identify clusters with similar densities but it is sensitive to the connected points (see Figure ~\ref{figconnect}).
In this research, our proposed VDPC incorporates DBSCAN's key strategies to complement the capability of DPC.

\subsection{Extensions of DPC}

Due to the reliable performance and the low complexity of DPC, there are many extended DPC models proposed in the literature. Most of these extensions improve the original DPC algorithm in two perspectives, namely similarity measurement and cluster formation strategy. Instead of geometric distance measures, feature learning methods may obtain better representations of the given dataset so as to achieve better clustering results.
Along this line of work, Li \emph{et al.} designed a new density measure based on tree structure \cite{li2018comparative}. Liu \emph{et al.} redefined DPC's parameters using SNN similarity \cite{liu2018shared}. Chen \emph{et al.} proposed an adaptive density clustering algorithm based on a new density measure, which finds density peaks in different density regions \cite{chen2019domain}.
Xu \emph{et al.} proposed to use density-sensitive similarity to identify manifold datasets \cite{xu2020robust}.
For identifying clusters with different density, Wang \emph{et al.} proposed IDPC, which computes a new relative density for individual data points and then takes a two-step strategy to obtain the final clusters \cite{wang2021relative}.
Many other extensions of DPC aimed to improve the clustering formation strategy striving for better clustering results.
Liu \emph{et al.} proposed the ADPC-KNN algorithm by autonomously merging initial clusters if they are density-reachable \cite{yaohui2017adaptive}.	
Hou \emph{et al.}  introduced a new concept of relative density relationship to identify cluster centers \cite{hou2020density}.
Fang \emph{et al.} developed a within-cluster similarity-based core fusion strategy \cite{fang2020adaptive}.
Du \emph{et al.} proposed an improved DPC algorithm based on KNN and PCA (Principal Component Analysis) to solve the issue that DPC may neglect certain clusters and obtain inferior results on high dimensional data \cite{DU2016135}. 
Xie \emph{et al.} proposed an improved DPC algorithm based on the fuzzy weighted KNN \cite{XIE201619}.
Chen \emph{et al.} proposed a novel clustering algorithm named CLUB, which finds the  density backbone of clusters on the basis of KNN and SNN \cite{Chen2016Effectively}.
Hou \emph{et al.} proposed density normalization to relieve the influence of the local density criterion \cite{hou2019enhancing}.
Mohamed \emph{et al.} proposed a KNN-based DPC using the SNN to identify dense regions \cite{abbas2021denmune}.
Chen \emph{et al.} replaced density peak with density core, which consists of multiple peaks and still retains the shape of clusters \cite{chen2018decentralized}.
To identify clusters with different density, Xie \emph{et al.} used density-ratio to replace $d_c$ (see (\ref{rho})) \cite{xie2018density}.

Nonetheless, all the afore-reviewed studies did not consider to improve DPC from the perspective of conducting an overview analysis on the clusters' property based on the data distribution. 
In this paper, we propose a systematic and autonomous analysis method to distinguish the different density levels exist in the given dataset, which may have variational density distributions patterns. Thus, we transform the complex variational density clustering problem into relatively straightforward uniform density clustering problems in the respective density levels. 
In the following section, we present the technical details of VDPC.

\section{Variational Density Peak Clustering (VDPC) Algorithm}
\label{vdpc}
In this research, we propose a novel Variational Density Peak Clustering (VDPC) algorithm, which consists of the following three key steps: (\rmnum{1})~representatives selection and initial clusters formation, (\rmnum{2})~dividing the initial cluster centers into density levels by systematically analyzing the distribution of data points, and (\rmnum{3})~final cluster formation. These three key steps are introduced in details in the subsequent subsections.
\subsection{Representatives Selection and Initial Cluster Formation }
\label{rsicf}

Representatives, denoted as $r$, are often selected to find the backbone of the given data points \cite{huang2019ultra}, which provides us the insights on the data distribution. The typical way to select representatives is through clustering, i.e., the identified cluster centers are often the representatives that we want to obtain. In VDPC, we first extend DPC to find the representatives. Specifically, we propose a parameter on the cut-off value of $\delta$, denoted as $\delta_t$, to select the representatives on the 2D decision graph (see Figure ~\ref{figoutliers}). Please note that although we use the same decision graph of DPC to determine the initial cluster centers in VDPC, we do not set the cut-off value of local density (e.g., $\rho_u$ used in DPC). Therefore, VDPC does not require any human intervention to determine the cluster centers.

\noindent {\bfseries Definition 1 (Representatives (\textit{r}))}. 
The data points with higher $\delta$ values are selected as representatives as follows: 
\begin{equation} \label{r}
r=\{x_i|\delta_i \geq \delta_t\}.
\end{equation}
 
\begin{figure}[!t]
	\centering
	\subfigure[]{
		\label{Fig5a}
		\includegraphics[width=0.4\textwidth]{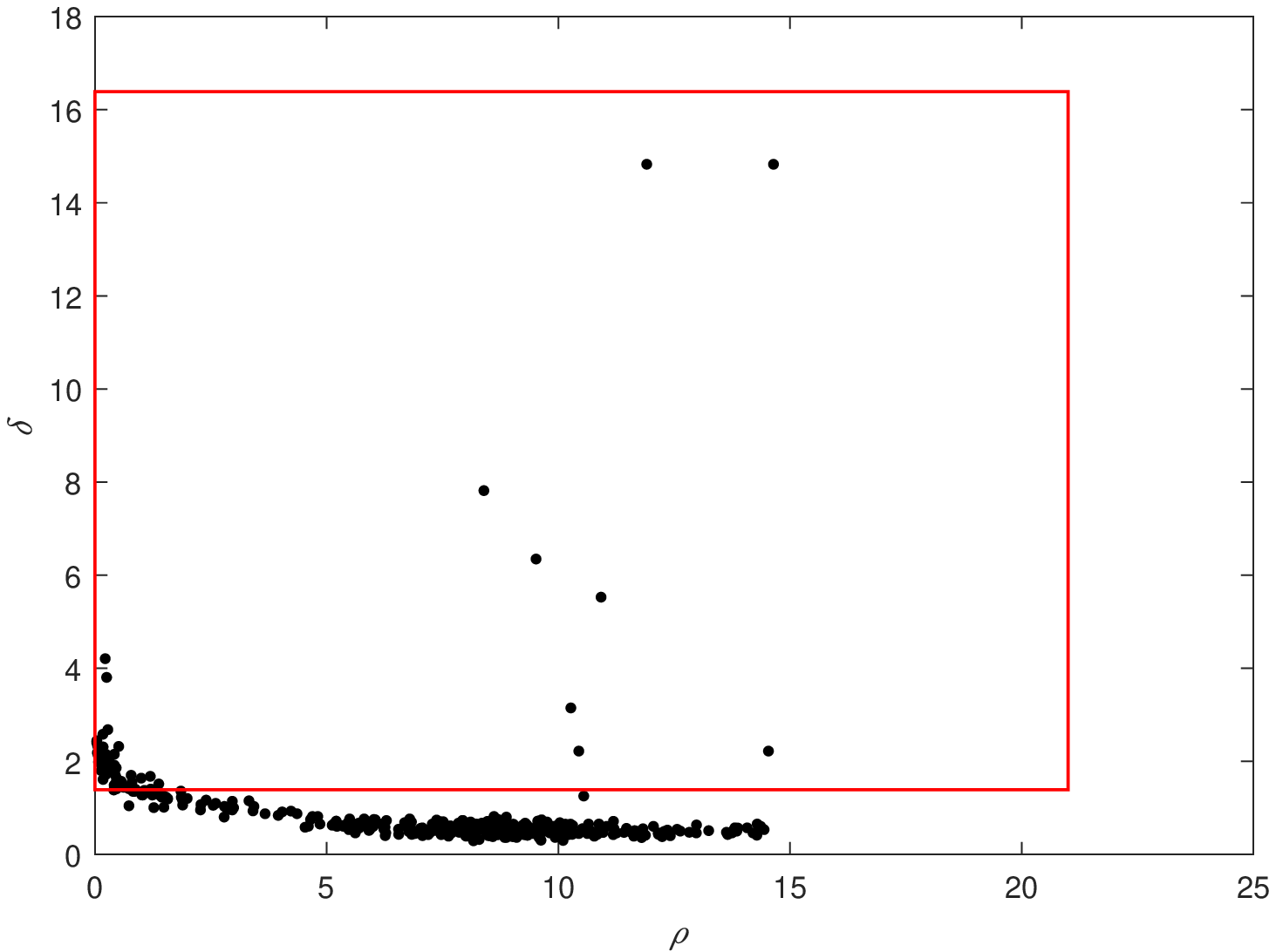}}
	\subfigure[]{
		\label{Fig5b}
		\includegraphics[width=0.4\textwidth]{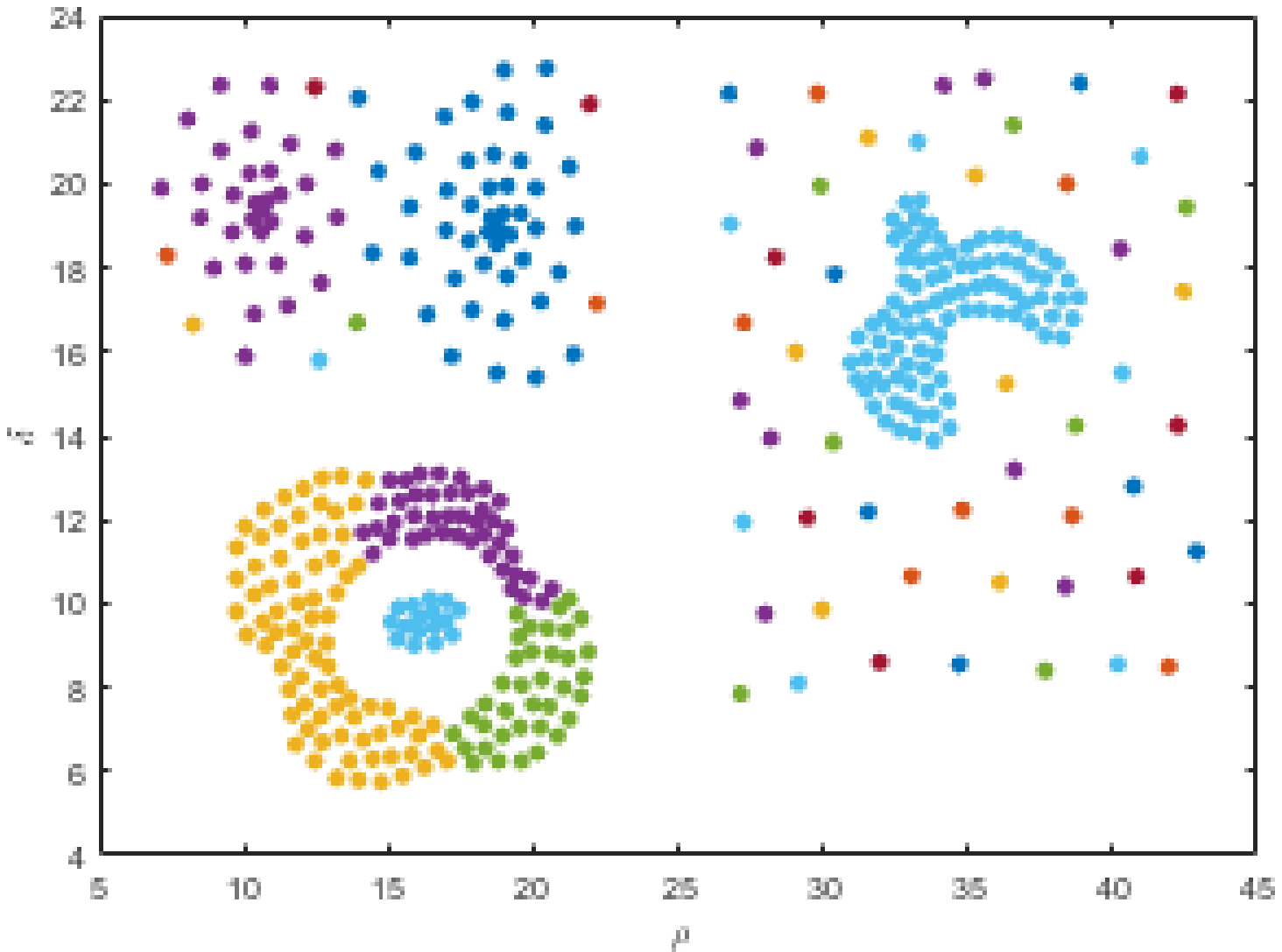}}	
	\caption{Representatives selection and initial cluster formation. (a) The representatives of dataset Compound \cite{Chang2008Robust}  (identified in the red rectangle) ($\delta_t=1.39$). (b) The initial clusters of dataset Compound are obtained by applying the clustering formation strategy of DPC on all the identified representatives. }
	\label{Fig3}
\end{figure}

As illustrated in Figure~\ref{Fig5a}, for dataset Compound, when $\delta_t$ is set to 1.39 (determined by heuristically fine-turing), the representatives in the red rectangle are selected. If we take these representatives as the initial cluster centers. The corresponding clusters obtained by applying DPC are shown in Figure~\ref{Fig5b} for illustration. Note that for a representative $x_i$, we denote the corresponding initial  cluster centered on $x_i$ as $C_{x_i}$.

Note that in this initial step, we consider all data points fulfilling selection criterion of $\delta_i \geq \delta_t$ as representatives regardless of their $\rho$ values. The key reason we deem $r$ as appropriate representatives (see (\ref{r})) is as follows: If a representative (cluster center) has a higher $\delta$ value, it is far away from other centers with higher local density according to the definition of parameter $\delta$ (see (\ref{delta})). As shown in Figure~\ref{Fig5b}, these selected representatives can be used to form initial clusters (assign all the data points to their corresponding representatives like DPC) and the distribution of these initial clusters helps us to understand the intrinsic structure of the underlying dataset.

\subsection{Dividing Representatives into Different Density Levels}
\label{levels}
After identifying the representatives and their corresponding initial clusters, we analyze their local density ($\rho$ in DPC) to better understand the intrinsic structure of the underlying dataset. This step mainly aims to divide the whole dataset (including the identified representatives) into a number of levels, wherein each level only comprises data points of similar density.

\begin{figure}[!t]	
	\centering
	\includegraphics[scale=0.7]{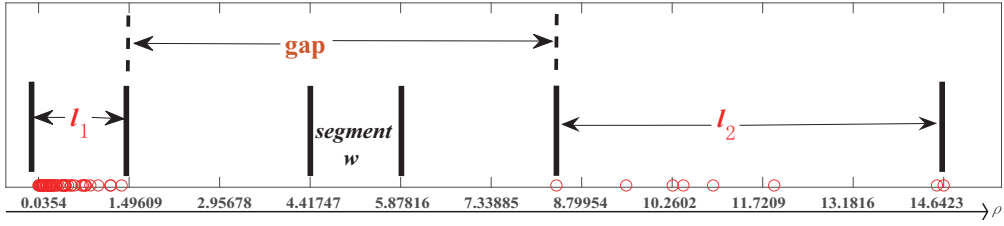}
	\caption{The density levels are determined by projecting the identified representatives onto the $\rho$-axis. The red circles represent the representatives and each segment has an equal  width of \textit{w}.
		For the Compound dataset, $w= 1.4607$ (see (\ref{w})) and the number of density levels $numl=2$.}	
	\label{figmap}
\end{figure}

Specifically, we first project the identified representatives from the 2D decision graph onto the $\rho$-axis (see Figure~\ref{figmap}), i.e., we tentatively ignore the $\delta_i$ values and only consider the $\rho_i$ values for now.
We then divide the $[min(\rho),max(\rho)]$ range into $num$ equal width segments. As such, the width $w$ of each segment can be computed as follows:
\begin{equation} \label{w}
w=\frac{max(\rho)-min(\rho)}{num},
\end{equation}
where $max(\rho)$ and $min(\rho)$ denote the largest and the smallest $\rho$ values  of all the representatives, respectively. Because when we conduct all experiments in this research work, we always set \textit{num} to a constant value, we do not consider it as a parameter that needs to be predefined by users.
The heuristic setting of this parameter \textit{num} is discussed in Section \ref{num}. Furthermore, we present the definitions of gap and local density levels, which are two important terms used in VDPC, as follows.

\noindent {\bfseries Definition 2 (gap (denoted as $gap$))}. 
For the identified representatives projected onto the $\rho$-axis, if the distance between any pair of the adjacent representatives is at least twice larger than the width ($w$) of a segment, we deem there is a gap in between and the width of the gap, denoted as \textit{gap}, is defined as the distance between the two adjacent representatives. Mathematically, we have the following definition:

\begin{equation} \label{gap}
gap_i=
\begin{cases}
\rho_{i+1}-\rho_{i}, & \textit{if} \quad\rho_{i+1}-\rho_i \geq 2\cdot w, \forall i \in [0, N-1], \rho_{i+1} > \rho_i,\\
 0,& otherwise.
\end{cases}
\end{equation}

If $\forall$$gap_i = 0$, we say there is no gap in the underlying dataset. Otherwise, we say there are $|gap_i > 0|$ number of gaps in the underlying dataset.
As illustrated in Figure~\ref{figmap}, the range of [1.375, 8.392] is the only identified gap in this case.

\noindent {\bfseries Definition 3 (density levels (denoted as $l$))}. The identified \textit{gap}(s) divide(s) the overall range ($\rho$) of representatives into \textit{numl}=$|gap_i>0|+1$ intervals. Moreover, we denote the identified intervals as $density$ $levels$ (denoted as $l$). 
 
As illustrated in Figure~\ref{figmap}, the overall range of representatives is $[0.0354,14.6423]$, and the range of the only gap is [1.375, 8.392] in this case (see (\ref{gap})). Obviously, this gap divides the representatives into two different intervals (i.e., density levels): lower density level $[0.0354,1.375]$ and higher density level $[8.392,14.6423]$.
Generally speaking, the difference between two different density levels is large. For example, with reference to Figure~\ref{figmap}, the average local density of all the representatives in the two density levels is computed as $\overline{\rho_{l_1}}=0.3685$ and $\overline{\rho_{l_2}}=11.3278$, respectively. The relative difference in terms of ratio $\overline{\rho_{l_2}}/\overline{\rho_{l_1}}\approx 31 $ is large. As a result, Definition 2 on the identification of whether there is any gap exists in the underlying dataset effectively helps VDPC to systematically determine whether there is significant difference among all the data points in terms of their local density.

Note that all the representatives are now divided into one or more density levels and all the data points are already assigned to different representatives to form initial clusters. Thus, 
all the initial clusters also fall into one or more density levels. 
In the following subsection, we introduce how the data points may be reassigned according to the identified number of density levels so as to accurately identify clusters with variational density.
\subsection{Final Cluster Formation}

After dividing all the representatives into $numl$ density levels,
we can further regulate the subsequent cluster formation process based on the value of \textit{numl}. 
Specifically, we apply different cluster formation processes for $numl=1$ and $numl\geq2$ correspondingly. When $numl = 1$, there is no gap found among the representatives, meaning all the clusters have similar density that falls on the same level. When $numl\geq2$, the clusters have significantly different (i.e., variational) density; therefore, further investigations are needed in this scenario. 
We deem a dataset has similar density if the corresponding \textit{numl} is found as 1. On the other hand, a dataset has variational density if $numl \geq 2$.
For example, as shown in Figure~\ref{figmap}, the Compound dataset has two density levels, $numl=2$, so it is a variational density dataset. We introduce the detailed cluster formation strategies in the following subsections.

\subsubsection{Cluster Formation for Datasets with Similar Density}

When $numl=1$, all the initial clusters fall into the same density level. In this situation, although all the initial clusters have similar density, some connected points may still exist  (see Figure~\ref{figconnect} as an example). Because DPC well handles the connected points that data points with lower $\rho$ values are assigned to the nearest higher density neighbors \cite{rodriguez2014clustering}, we simply adopt DPC for the subsequent cluster formation. Specifically, we take all the identified representatives as the centers of the final clusters, then each remaining data point is sequentially assigned to its nearest higher density neighbor so as to generate the final clusters. In this scenario, the clustering strategy of VDPC is identical to DPC.

Note that the VDPC does not straightforwardly replicate DPC in this scenario (only the clustering strategy is adopted as the same as DPC), because the initial cluster centers are systematically identified in VDPC whereas the cluster centers have to be manually selected in DPC.
\subsubsection{Cluster Formation for Datasets with Variational Density }
\label{num2}
When $numl\geq2$, initial clusters fall into different density levels with large gap(s) in between.
In this scenario, the cluster formation strategy in different density levels should be different so as to effectively cope with the different data distribution patterns in the respective density levels. Specifically, we identify clusters with lower density in $l_1$, and we propose an autonomous DBSCAN algorithm to group the remaining data points with uniform density in the $p$th density level except $l_1$ ($l_p, 2\leq  p\leq numl$).

In $l_1$, there are two kinds of data points and they all have lower local density. 
The first kind are the boundary points locating at the boundary of clusters with higher density in $l_p$, and the other kind are the data points in the clusters with lower density  (see Figure \ref{lowhigh}).
Boundary points are supposed to be a part of clusters with higher density, so we try to delete them from $l_1$ and reassign them to the nearest representatives in $l_p$.
To this purpose, we propose an autonomous Shared Nearest Neighbour Clustering (aSNNC) algorithm to group all the data points in $l_1$ into clusters. SNNC (see Section \ref{mknn}) requires users to predetermine the number of nearest neighbours $k$ \cite{patidar2012analysis}. In VDPC, we set a global $k$ value of SNNC in an autonomous way by heuristically setting $k=\lceil \sqrt{nl} \rceil$, where $nl$ denotes the number of data points fall into $l_1$.
Please refer to Appendix \ref{assnnc} for more details on setting the heuristic value of \textit{k}. 


For an initial cluster, from its center to its boundary, the local density of data points are getting lower and lower. 
The number of boundary points in $l_1$ is always small because most data points have been merged into initial clusters located in $l_p$  (in most real-world datasets, there are much lesser number of outliers than the number of data points in a dense cluster, see Section \ref{uci}). Based on this nature of boundary points, we give the definition of boundary points in $l_1$  as follows.

\noindent {\bfseries Definition 4 (boundary points in $l_1$  (\textit{BP}))}.
For the clusters generated by applying aSNNC in $l_1$, denoted as $C_{l_1}=\{c^1_{l_1},\dots, c^i_{l_1},\dots,c^n_{l_1}\}$, where $n$ denotes the total number of clusters generated in $l_1$, if the number of data points in one cluster is smaller than the mean number of data points in $C_{l_1}$, the corresponding cluster may consist of boundary points. As such, the boundary points (\textit{BP}) are identified as follows:
\begin{equation} \label{b}
 \textit{BP}=\{\textit{all data points in}\textit{ }c^i_{l_1}|\#c_i<\frac{\sum_{i=1}^n\#c_i}{n}\},
\end{equation}
where $c^i_{l_1}$ denotes a cluster generated by applying aSNNC in $l_1$, $\#c_i$ denotes the number of data points in cluster $c^i_{l_1}$.

The remaining clusters in $l_1$ are the final clusters with lower density, denoted as $C_{low}$:
\begin{equation} \label{clow}
C_{low}=\{  c^i_{l_1} |\#c_i\geq\frac{\sum_{i=1}^n\#c_i}{n}\}.
\end{equation}

After identifying the clusters with lower density ($C_{low}$) in $l_1$, we reassign the identified boundary points (\textit{BP}) to the nearest representatives in $l_p$. The data points $\{x_i|x_i \in D$ and $x_i \notin  C_{low}\}$, denoted as $\vert D \vert-\vert C_{low} \vert$, have similar density in the respective density level $l_p$, so we further employ DBSCAN to identify the remaining clusters. As afore-introduced in Section \ref{dbscan}, the performance of DBSCAN highly depends on and is sensitive to the values of the two parameters: \textit{Eps} and \textit{MinPts}.
In this research, we propose an autonomous DBSCAN (aDBSCAN) algorithm to systematically determine the values of the two parameters only based on the data distribution patterns observed in $l_p$.

\begin{figure}[!t]	
	\centering
	\includegraphics[scale=0.4]{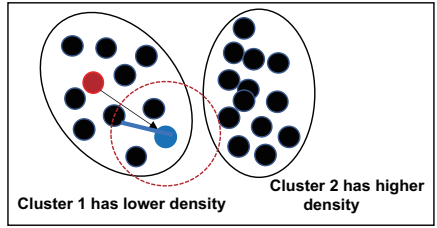}
	\caption{Assume both clusters are found in density level $l_p$. The red point is identified as $x_{low}$, the blue point is identified as $x_{\textit{far}}$, then the length of the blue segment is set to \textit{Eps}.}
	\label{figeps}
\end{figure}

The parameter \textit{Eps} of DBSCAN determines whether data points are directly density-reachable from core points or not. In order to avoid boundary points previously reassigned to $l_p$ being identified as noises by DBSCAN, 
the value of \textit{Eps} should be set carefully to let them have a greater possibility of being density-reachable from core points. Thus, 
the parameter \textit{Eps} is heuristically defined as follows:

\begin{equation} \label{eps}
\textit{Eps}=sim(\lceil \sqrt{\vert C_{x_{\textit{low}}} \vert} \rceil ),
\end{equation}

\begin{equation} \label{eps2}
sim(i)=\left \lVert x_{\textit{far}}-x_i \right\rVert, x_i \in l_p,
\end{equation}
where \textit{sim} denotes a vector of Euclidean distance sorted in an ascending order between data point $x_{\textit{far}}$ and any other data points in $l_p$, 
$x_{\textit{far}}$ denotes the farthest boundary point in the initial cluster $C_{x_{\textit{low}}}$ from  its representative $x_{\textit{low}}$ with the lowest local density in $l_p$,
$\vert C_{x_{low}} \vert$ denotes the number of data points in $C_{x_{\textit{low}}}$, 
and $(\cdot)$ denotes the index of the vector \textit{sim}. We use $x_{\textit{far}}$ to compute \textit{sim} so that at least  $\lceil \sqrt{\vert C_{x_{\textit{low}}} \vert} \rceil$ data points are within distance $Eps$ from $x_{\textit{far}}$, as a result, the boundary points have a greater possibility of being density-reachable from core points.
An example is shown in Figure~\ref{figeps} to depict the computing procedure of $\textit{Eps}$.
Please refer to Appendix \ref{adbscan} for more details on the heuristic setting of the cut-off value $\lceil \sqrt{\vert C_{x_{\textit{low}}} \vert} \rceil $.

After obtaining the value of \textit{Eps}, we subsequently use it to systematically determine the value of \textit{MinPts}. If the value of \textit{MinPts} is set too high, the boundary points may be highly likely identified as noises; if the value of \textit{MinPts} is set too low, the boundary points may be identified as core points. In order to avoid these extreme cases from happening, we take a balanced trade-off. 
Specifically,
we first obtain the number of data points that have a shorter distance to the representative with the highest local density in $l_p$ than \textit{Eps} as follows:
\begin{equation} \label{test}
\textit{MinPts}_{\textit{high}}=\sum_{j}\chi(\lVert x_{\textit{high}},x_j\rVert-\textit{Eps}), x_{\textit{high}}\in C_{x_{\textit{high}}},x_j \in C_{x_{\textit{high}}},
\end{equation}
\begin{equation} \label{test}
\chi(v)=
\begin{cases}
1, & v<0,\\
0, & v\geq0,
\end{cases}
\end{equation}
where $x_{high}$ denotes the representative with the highest local density in $l_p$. Similarly, 
we then obtain the number of data points that have a shorter distance to $x_{\textit{far}}$ (see (\ref{eps2})) than \textit{Eps} as follows:

\begin{equation} \label{minptsspa}
 \textit{MinPts}_{\textit{low}}=\sum_{j}\chi(\lVert x_{far},x_j \rVert-\textit{Eps}), x_{\textit{far}}\in C_{x_{\textit{low}}},x_j \in C_{x_{\textit{low}}}.
\end{equation}

To strive for a balanced, reasonable trade-off, 
we set the average of $\textit{MinPts}_{\textit{low}}$ and $\textit{MinPts}_{\textit{high}}$ as the value of \textit{MinPts} in aDBSCAN  as follows. 
\begin{equation} \label{MinPts}
\textit{MinPts}=\lceil \frac{\textit{MinPts}_{\textit{low}}+\textit{MinPts}_{\textit{high}}}{2}\rceil.
\end{equation}

\begin{figure}[!t]	
	\centering
	\includegraphics[scale=0.5]{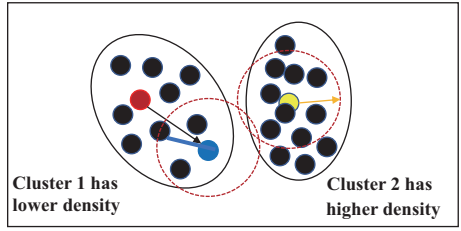}
	\caption{By extending Figure~\ref{figeps}, where the blue point is identified as  $x_{\textit{far}}$, we further identify the yellow point as $x_{\textit{high}}$. The value of \textit{MinPts} is then computed as the average of the numbers of data points within the respective \textit{Eps}  radius, i.e., $\textit{MinPts}=\lceil(3+9)/2\rceil= 6$.}
	\label{figminpts}
\end{figure}

\noindent An example is shown in Figure~\ref{figminpts} to illustrate the determination process of \textit{MinPts}.


So far, we have shown that the parameter values in aDBSCAN are both autonomously determined (see (\ref{eps}) and (\ref{MinPts}), respectively). By applying aDBSCAN with the auto-obtained parameter values, the data points in the respective density level ($l_p$) can be divided into either belong to the formed clusters or belong to the set of noises. 
Let $\textit{NCD}$ denote the set of all data points in the formed clusters and let \textit{CO} denote the set of all data points identified as noises, then we use $|\textit{NCD}|$ and $ |\textit{CO}|$ to denote the number of clusters in \textit{NCD} and the number of data points in  \textit{CO}, respectively. Then, we further improve the clustering results of aDBSCAN by reexamining the micro-clusters in \textit{NCD}.

\noindent {\bfseries Definition 5 (micro-clusters in \textit{NCD})}. For all clusters in \textit{NCD}, if the local density of a cluster center is smaller than the averaged local density of all centers, we call its corresponding cluster a micro-cluster (denoted as \textit{mc}) and we use $|\textit{mc}|$ to denote the number of micro-clusters.

It is not guaranteed that all the micro-clusters obtained in $l_p$ constitute the final clusters. Based on Definition 5, the centers of the identified micro-clusters are already found as having relatively lower $\rho$ values, we need to further check whether they could be considered as the final cluster centers. To make such a decision, we apply a straightforward heuristic rule to check whether an unnecessary number of clusters have been identified in $l_p$. The heuristic rule and the corresponding action are defined as follows:
If \textit{$|mc|$}\textit{$<\frac{|\textit{NCD}|}{2}$}, it means the micro-clusters not only have smaller $\rho$ values, but also are small in number, then all the data points in \textit{mc} are merged into other nearest clusters (non-micro-clusters) in \textit{NCD}; 
otherwise, i.e., \textit{$|\textit{mc}|$}\textit{$\geq\frac{|\textit{NCD}|}{2}$}, it means the micro-clusters constitute the majority cases of the identified $|\textit{NCD}|$ clusters, then the micro-clusters are deemed as the final clusters.
After the identification of the final cluster centers, the remaining data points in \textit{CO} are sequentially assigned to the nearest cluster center of higher local density (see DPC's assignment process in Section~\ref{dpc}).

For both kinds of different density levels $l_1$ and $l_p$, we apply two different clustering methods: aSNNC and aDBSCAN (see Section~\ref{num2}). In order to verify the effectiveness of such identification pipeline, we conduct a heuristic validation by alternately applying aDBSCAN and aSNNC in $l_1$ and $l_p$. The results (see Appendix~\ref{dbandsnnc}) show that the current configuration on applying aSNNC in $l_1$ and aDBSCAN in $l_p$ leads to the best performance, which suggests that our intuitive designs of aSNNC and aDBSCAN are appropriate.

\subsection{Overall VDPC Procedures}

\begin{figure}[!t]	
	\centering 
	\includegraphics[scale=0.7]{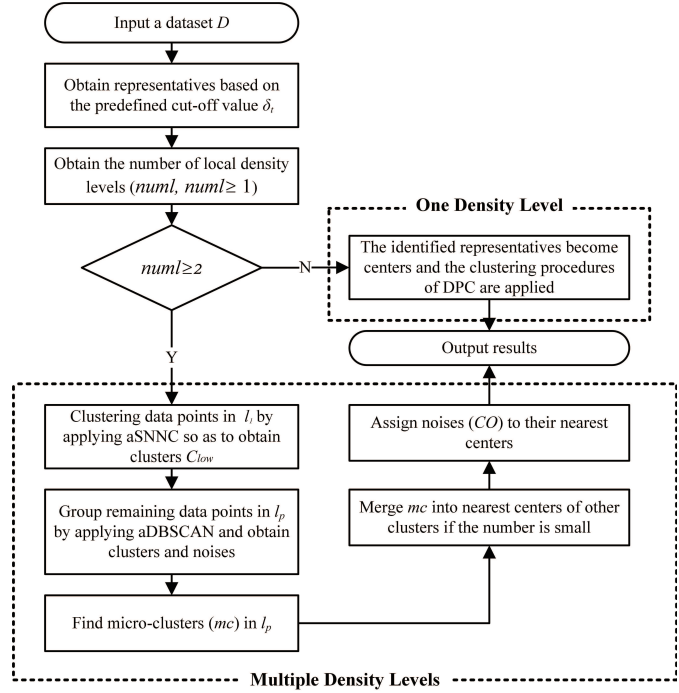}
	\caption{The flowchart of VDPC.}
	\label{figflow}
\end{figure}

\begin{figure}[H]
	\centering
	\includegraphics[scale=0.45]{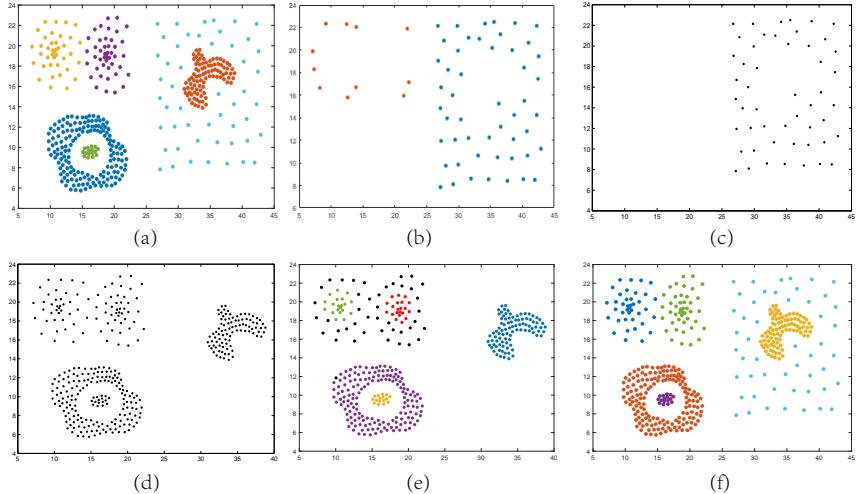}				
	\caption{Step-by-step illustrations of applying VDPC on the Compound dataset. (a)~The ground-truth clusters given in the Compound dataset \cite{Zahn1971Graph}. (b)~The clustering results on data points fall in $l_1$ by applying aSNNC. (c)~The final cluster ($ C_{low} $) identified in $l_1$. The data points in the top-left region are not identified as final clusters because they are identified as boundary points. (d)~The remaining $ \left| D \right| -\left| C_{low} \right|$ data points that require further analysis. (e)~Results of applying aDBSCAN on the remaining $ \left| D \right| -\left|C_{low} \right|$ data points, the black points represents noises, and the red, green and orange clusters are three identified micro-clusters. Note that the micro-clusters in this case are the final clusters because their size is large (see Section(\ref{num2})). (f)~Final clustering results after assigning noises and micro-clusters.}
	\label{figexcom}
\end{figure}

Till now, we have introduced all the procedures of VDPC in the preceding subsections and we summary the overall algorithm in this subsection. Notably, as an extension to DPC, VDPC only requires two predefined parameter values, which is one number less than that required by DPC, namely \textit{pct}  (adopted from DPC, see (\ref{rho})) and $\delta_t$. In VDPC, the processes of identifying the cluster centers and subsequent cluster assignment are all performed in a systematic way without the need to predetermine other parameter values. Comparing to DPC, wherein human intervention on the identification of cluster centers is required, VDPC is definitely autonomous throughout the overall clustering procedures.
The overall VDPC algorithm is shown in Algorithm \ref{alg:all} and the corresponding flowchart is shown in Figure~\ref{figflow}. To provide a better understanding of the VDPC procedures, we present the step-by-step illustrations in Figure~\ref{figexcom}.

\begin{algorithm}[H]
	\KwIn{ dataset $D$, parameters $pct$ and $\delta_t$}
	\KwOut{assigned cluster indices for all data points in $D$}
	obtain representatives according to $pct$ and $\delta_t$  (see Section \ref{rsicf}) \;
    obtain the number of different density levels 	$numl$ (see Definition 3)\;	
	\eIf{$numl \geq2$}{
		 clusters $C_{low}$$\leftarrow$ clustering the data points in $l_1$ by applying aSNNC with the boundary points excluded (see (\ref{b}))\; 			
	 clusters and noises $\leftarrow$ clustering the remaining $\vert D \vert-\vert C_{low} \vert$ data points by applying aDBSCAN\;				
			micro-clusters ($mc$) $\leftarrow$ clusters in $l_p$ that the local density of their center is smaller than the averaged local density of all centers (see Definition 5)\;
            $|mc|$ $\leftarrow$ the number of micro-cluster (\textit{mc}) \;
			\If{$|mc|<|\textit{NCD}|/2$}{					
				assign all data points in the  micro-clusters to the nearest centers of other clusters\;
			}				
			assign noises (\textit{CO}) to their nearest centers (see DPC's assignment process in Section \ref{dpc})\;				
	}{
		take the identified representatives as centers and merge other data points into their nearest higher density centers\;
	}
	output the final clustering results\;
	\BlankLine
	\caption{Variational density peak clustering algorithm (VDPC)}
	\label{alg:all}
\end{algorithm}

\subsection{Analysis of Time Complexity}
In VDPC, the time complexity of similarities matrix computation (see (\ref{educa})) is $O(n^2)$. For the scenario of cluster formation for datasets with non-variational densities, the time complexity is the same as that of DPC, which is $O(n^2)$. For the scenario of cluster formation for variational density datasets, the time complexity of both aSNNC and aDBSCAN is $O(n^2)$. Thus, the overall time complexity of VDPC is on the same magnitude as DPC and DBSCAN, which is $O(n^2)$.

\section{Experimental Results and Discussions}
\label{erad}
In this section, we use eight synthetic datasets, six UCI datasets and six image datasets to comprehensively evaluate the effectiveness of the proposed VDPC algorithm. Datasets Jain, Flame, Aggregation, R15, Compound, Pathbased are from University of Eastern Finland\footnote{http://cs.uef.fi/sipu/datasets/}, large-scale datasets T58 and T710  are from Karypis Lab\footnote{http://glaros.dtc.umn.edu/} and they have no groundtruth defined. Datasets German, Yeast, Pima, Heart, Spambase and Immunotherapy are from UCI datasets\footnote{https://archive.ics.uci.edu/ml/index.php}. 
COIL-20, Olivetti Faces-100, Olivetti Faces, UMist Faces, mini-Corel5K and mini-Cafar10 consist of faces and general object images, respectively. The statistical information of all the datasets used in this paper are shown in Table \ref{features}. All the datasets used for experiments and the programming codes of VDPC can be downloaded from our github  repository\footnote{https://github.com/mlyizhang/Datasets.git}.

\begin{table}[htbp] 
	\centering   
	\caption{ Dataset Features}  
	\label{features}  
	\setlength{\tabcolsep}{0.5mm}{
		\begin{tabular}{cccccc}  
			\toprule  
	Type&ID	&	Datasets&	\#Samples&	\#Dimensions&	\#Natural clusters\\
			\midrule  
		Synthetic	&1&Jain \cite{Chang2008Robust} &373&2&2\\ 
		Synthetic	&2&Flame \cite{fu2007flame}&240&2&2\\
		Synthetic	&3&Aggregation \cite{Givoni2009A} &788&2&7\\ 	
		Synthetic	&4&R15 \cite{Veenman2002A} &600&2&2\\
		Synthetic	&5&Compound \cite{Zahn1971Graph} &399&2&6\\ 
		Synthetic	&6&Pathbased \cite{Chang2008Robust} &300&2&2\\		
	Synthetic	&7	&T58&8000&2&$N/A$\\
	Synthetic	&8	&T710&10000&2&$N/A$\\
	\hline
	UCI	&1&	German&1000&24&2\\
		 UCI	&2&Yeast&1484&8&3\\
		 	 UCI	&3&	Pima&768&8&2\\
	 UCI	&4&	Heart&300&13&2\\
	 UCI	&5&	Spambase&4601&57&2\\
	 UCI	&6&	Immunotherapy&90&7&12\\
		\hline
	  Images	&1&COIL-20&1440&128x128&20\\
	 Images&2&	Olivetti Faces-100	&100&64x64&10\\
	  	Images&3&	Olivetti Faces&400&64x64&40\\
	Images	&4&	 UMist Faces&575& 112x92 &20\\
	  Images	&5&	mini-Corel5K&100&192x128&10\\
	Images	&6& mini-Cafar10&1000&32x32&10\\
			\bottomrule  
	\end{tabular} }
\end{table} 

We present the selected $pct$ and $\delta_t$ values for all the datasets in the respective experiments. In order to fairly compare the performances of all algorithms, we run all algorithms using the same machine (i9-10900K CPU and RAM 32 GB) and software (MATLAB R2020a) to obtain the clustering results.

\subsection{Sensitivity Test on Parameter \textit{num}}
\label{num}

The proposed VDPC requires two user-defined parameters ($pct$ and $\delta_t$). It is because when conducting experiments, we set \textit{num} 
to a constant value, i.e., 10, according to the sensitivity test.
Specifically, we select six datasets Jain, Flame, Aggregation, R15, Compound, and Pathbased in this sensitivity test because they vary in sizes (from 240 to 788), classes (from 2 to 7), and different density levels (1 or 2).
As shown in Table \ref{tablenum}, when we take different values of \textit{num} (namely 5, 8, 10, 12, 15) , VDPC always obtains the best performance when $num = 10$. Moreover, when we set \textit{num}=10, datasets Jain, Flame, Aggregation, R15 are found having one density level, while the rest having two density levels. This finding suggests that by setting num to 10, different density levels can be effectively identified from different datasets while VDPC still obtains the best performance. Thus, for all datasets used in this paper, we always set \textit{num}=10 in VDPC.

\begin{table}[htbp] 
	\centering   
	\caption{ Clustering results (ARI) of VDPC using different $num$ values}  
	\label{tablenum}  
	\setlength{\tabcolsep}{0.5mm}{
		\begin{tabular}{ccccccccccc}  
			\toprule  
			\multirow{3}{*}{Datasets}
			&\multicolumn{5}{c}{$num$}&\\	
			\cline{2-7}
			&5&8&10&12&15\\
			\midrule
			Jain&1.0000&1.0000&1.0000&1.0000&1.0000\\
			Flame&1.0000&1.0000&1.0000&1.0000&1.0000\\
			Aggregation&1.0000&1.0000&1.0000&1.0000&1.0000\\		
			R15&1.0000&1.0000&1.0000&1.0000&1.0000\\
			Compound&0.4867&1.0000&1.0000&1.0000&0.9859 \\
			Pathbased&1.0000&1.0000&1.0000&0.5003&0.5422\\
			\bottomrule  
	\end{tabular} }
\end{table}

\subsection{Benchmarking Models}
To show the effectiveness of the proposed VDPC algorithm, we select two classical clustering (including DBSCAN \cite{DBSCAN} and original DPC \cite{rodriguez2014clustering}) and four state-of-the-art improved DPC algorithms (including DPC-KNN\footnote{https://github.com/mlyizhang/DPC-KNN-PCA.git} \cite{DU2016135}  , McDPC\footnote{https://github.com/mlyizhang/Multi-center-DPC.git} \cite{wang2020mcdpc}, SNNDPC\footnote{https://github.com/liurui39660/SNNDPC.git} \cite{liu2018shared} and FKNN-DPC\footnote{https://github.com/liurui39660/SNNDPC.git} \cite{XIE201619}. The parameters used in each algorithm are introduced in Table \ref{tabledes}. For fair comparisons, all the parameters values are fine-turned with an ample number of experiments and the respectively selected best performing parameter values are listed in Tables \ref{tablesyn}-\ref{tableimage}.
To quantitatively compare the performance of all algorithms, we use two popular metrics, namely Adjusted Rand Index (ARI) \cite{vinh2010information} and Normalized Shared Information (NMI) \cite{estevez2009normalized}. 
\begin{table}[htbp] 
	\centering
	\small 
	\caption{ The parameter description of each algorithm}  
	\label{tabledes}  	
	\begin{tabular}{lp{10cm}lp{8cm}}  
		\toprule 	 
		Algorithm&	Required parameters with descriptions\\ 
		\midrule  
		DBSCAN &\textit{Eps}: radius of reachable neighborhood\\	
		&\textit{MinPts}: min number of points within radius \textit{Eps}  to form a cluster\\	
		\hline	
		DPC&$pct$:  a relative ratio to determine the cut-off distance $d_c$ \\
		&\textit{$\rho_u$}: determined by users to select centers  in the decision graph\\
		&\textit{$\delta_u$}: determined by users to select centers  in the decision graph\\	\hline
		DPC-KNN &$pct$: a relative ratio to determine the cut-off distance $d_c$\\
		&$d$: number of principal components to be selected after applying PCA.\\
		&$k$: number of nearest neighbors.\\
		\hline
		McDPC &$\gamma$: parameter used to perform $\rho$-cut\\		
		&$\theta$: threshold used to perform $\delta$-cut \\
		&$\lambda$:  threshold used to identify micro-clusters \\
		&$pct$:   a relative ratio to determine the cut-off distance $d_c$\\
		\hline
		SNNDPC&$k$: number of nearest neighbors\\
		&$NC$: number of centers\\
		\hline
		FKNN-DPC&$k$: number of nearest neighbors\\
		&$NC$: number of centers\\
		\hline
		VDPC  (ours)&$pct$: a relative ratio to determine the cut-off distance $d_c$\\
		&$\delta_t$: parameter used to select representatives\\
		\bottomrule  
	\end{tabular} 
\end{table} 
\subsection{Experiments on  Synthetic Datasets}
Table \ref{tablesyn} shows evaluation metrics ARI and NMI of the first six synthetic datasets because datasets T710 and T58 have no ground-truth labels. The corresponding algorithm parameter values are listed in 
Table \ref{tablesyn}.

According to our proposed intrinsic structure analysis method, datasets Flame, Aggregation, R15, Jain and T58 have one density level, while datasets Compound, Pathbased and T710 have two density levels. 
Obviously, whether it is one density level dataset or two density levels datasets, VDPC always achieves the best results especially for the challenging variational density datasets. DPC is only able to obtain the best results on one density level datasets. DBSCAN does not achieve the best results largely due to the influence of connected points. McDPC achieves the best results on all datasets expect Compound because it cannot identify clusters with variational density in the higher density level. The other three state-of-the-art extended DPC methods (namely DPC-KNN, SNN-DPC and FKNN-DPC) only achieve the best results on a small number of synthetic datasets that noticeably, they do not perform well on datasets with variational density.

\begin{table}[!t]
	\centering 
	\begin{threeparttable}
		\centering   
		\caption{ Comparison of six clustering algorithms on  synthetic datasets}  
		\label{tablesyn}  
		\small 
		\begin{tabular}{ccccc|cccc}  
			\toprule
			Algorithms &Par&Val&ARI&NMI&Val& ARI& NMI\\
			\midrule  
			\multicolumn{5}{c|}{\bf Dataset Flame}&\multicolumn{3}{c}{\bf Dataset Aggregation} \\					
			DBSCAN&Eps/MinPts&1/6&0.9280&0.8583&1.21/6&0.9828& 0.9749\\ 			
			DPC&$pct$&5&{\bfseries 1.0000}&{\bfseries 1.0000}&4&{\bfseries 1.0000}&{\bfseries 1.0000}\\
			DPC-KNN&$pct/d/k$&1/2/3&{\bfseries 1.0000}&{\bfseries 1.0000}&0.5/2/3&0.9957 &0.9884\\ 
			McDPC&$\gamma/\theta/\lambda/pct$&2/0.001/3/4&{\bfseries 1.0000}&{\bfseries 1.0000}&0.5/0.1/2.9/4&{\bfseries 1.0000} &{\bfseries 1.0000}\\
			SNNDPC&$k/NC$&5/2& 0.9502 &0.8994  &15/7&0.9594 &0.9555 \\ 
			FKNN-DPC&$k/NC$&6/2& {\bfseries 1.0000}&{\bfseries 1.0000} &20/7&0.7150  &0.8618  \\ 
			VDPC&$pct/\delta_t$&5/5.5&{\bfseries 1.0000}&{\bfseries 1.0000}&4/2.9&{\bfseries 1.0000}&{\bfseries 1.0000}\\ 
			\hline 		
			\multicolumn{5}{c|}{\bf Dataset R15}	&\multicolumn{3}{c}{\bf Dataset Compound}\\ 		
			
			DBSCAN&Eps/MinPts&0.3/3&0.9018&0.8942&1/5&0.9103&0.8774\\ 		
			DPC&$pct$&5&{\bfseries 0.9928}&{\bfseries  0.9942 }&5&0.6368&  0.5263\\
			DPC-KNN&$pct/d/k$&0.5/2/3&{\bfseries 0.9928}&{\bfseries  0.9942 }&5/2/8&0.5448 &0.7423\\
			McDPC&$\gamma/\theta/\lambda/pct$&1/0.001/1.02/0.1&0.9228 &0.9765&0.5/0.01/3/1&0.6074  &0.7781 \\  
			SNNDPC&$k/NC$&10/15& {\bfseries 0.9928}  &{\bfseries  0.9942 }   &4/6&0.8629 &0.9120  \\ 
			FKNN-DPC&$k/NC$&27/15&0.9892  &0.9913   &15/6&0.8229  &0.8362  \\  
			VDPC&$pct/\delta_t$&5/1&{\bfseries 0.9928}&{\bfseries  0.9942 }&1.9/1.39&{\bfseries 1.0000}&{\bfseries 1.0000}\\ 
			\hline 	
			\multicolumn{5}{c|}{\bf Dataset Jain}&\multicolumn{3}{c}{\bf Dataset Pathbased}\\ 
			DBSCAN&Eps/MinPts&2.9/20&{\bfseries 1.0000}&{\bfseries 1.0000}&1/4&0.6288&0.4577\\ 
			DPC&$pct$&40&{\bfseries 1.0000}&{\bfseries 1.0000} &5&0.6600&0.4572\\
			DPC-KNN&$pct/d/k$&3/2/15&0.5692 &0.5420&0.5/2/5&0.5448 &0.7423\\
			McDPC&$\gamma/\theta/\lambda/pct$&0.1/2/3.35/2&{\bfseries 1.0000}&{\bfseries 1.0000}&0.12/0.8/3.5/0.5&{\bfseries 1.0000}&{\bfseries 1.0000}\\   		
			SNNDPC&$k/NC$&12/2&{\bfseries 1.0000}  &{\bfseries 1.0000}  &9/3&0.9294  &0.9013\\  
			FKNN-DPC&$k/NC$&10/2&0.0562 &0.2330   &15/7&0.4729 &0.5783 \\ 
			VDPC&$pct/\delta_t$&50/5.5&{\bfseries 1.0000}&{\bfseries 1.0000}&0.4/3.5&{\bfseries 1.0000}&{\bfseries 1.0000}\\ 
			\bottomrule 
		\end{tabular}
The best results are highlighted in boldface.	  
	\end{threeparttable}

\end{table}

\subsection{Experiments on UCI Real-world Datasets}
\label{uci}
The clustering results of six UCI datasets, namely German, Yeast, Pima, Heart, Spambase and Immunotherapy, are shown in Table \ref{tableuci}. VDPC obtains the best ARI and NMI values for all datasets. The clustering results are encouraging because they suggest that VDPC has the ability of handling real-world datasets.

\begin{table}[htbp]
	\centering 
	\begin{threeparttable}
		\centering   
		\caption{ Comparison of six clustering algorithms on UCI datasets}  
		\label{tableuci}  
		\small 
		\begin{tabular}{ccccc|cccc}  
			\toprule
			Algorithms &Par&Val&ARI&NMI&Val& ARI& NMI\\
			\midrule  
			\multicolumn{5}{c|}{\bf Dataset German}&\multicolumn{3}{c}{\bf Dataset Yeast} \\					
			DBSCAN&Eps/MinPts&8.1/2&0.0810&0.0056&0.2/2&0.0031& 0.0157\\ 	
			DPC&$pct$&1&0.0042&0.0027&1&0.0014&0.0273\\
			DPC-KNN&$pct/d/k$&1/2/3&0.0542&0.0133&1/2/3&0.0034&0.0161\\ 
			McDPC&$\gamma/\theta/\lambda/pct$&2/0.01/14/4&0.0551&0.0436&2/0.3/0.1/1&0.0211 &0.0265\\
			SNNDPC&$k/NC$&25/2& 0.0531 &0.0146  &10/8&0.0142&0.0213  \\ 
			FKNN-DPC&$k/NC$&16/2&0.0446&0.0094  &7/8&0.0112  &0.0178    \\ 
			VDPC&$pct/\delta_t$&0.5/10&{\bfseries 0.0856}&{\bfseries 0.0807}&1/0.1&{\bfseries 0.0273}& {\bfseries0.2032} \\ 
			\hline 		
			\multicolumn{5}{c|}{\bf Dataset Pima}	&\multicolumn{3}{c}{\bf Dataset Heart}\\ 					
			DBSCAN&Eps/MinPts&0.5/2&0.0023&0.0042&0.8/3&0.0818  &0.1292  \\ 		
			DPC&$pct$&1&0.0218&0.0058&2&0.1723& 0.1537  \\
			DPC-KNN&$pct/d/k$&0.5/7/3&0.0131&0.0035&0.2/2/2&0.0628 &0.1180 \\
			McDPC&$\gamma/\theta/\lambda/pct$&0.02/1/0.6/5&0.1344 &0.0644 &2/1/1.6/0.2&0.1723 &0.1507\\  
			SNNDPC&$k/NC$&7/2& 0.0119  &0.0029   &22/2&0.1508  &0.1297  \\ 
			FKNN-DPC&$k/NC$&7/2&0.0453   &0.0175   &16/2&0.0715    &0.0860 \\  
			VDPC&$pct/\delta_t$&1/0.19&{\bfseries 0.1701}&{\bfseries 0.1636   }&0.01/1.6&{\bfseries0.1897  }&{\bfseries 0.1518 } \\ 
			\hline 	
			\multicolumn{5}{c|}{\bf Dataset Spambase}&\multicolumn{3}{c}{\bf Dataset Immunotherapy}\\ 
			DBSCAN&Eps/MinPts&9000/2&0.0000&0.0000&400/2&0.0000&0.0000\\ 
			DPC&$pct$&1&0.0643&0.0695    &1&0.0212 &0.1936 \\
			DPC-KNN&$pct/d/k$&1/3/2&0.0155 &0.0742&0.5/2/5&0.0046  &0.2906 \\
			McDPC&$\gamma/\theta/\lambda/pct$&3/2/10/2&0.1296 &0.0801 &1/1/25/1&0.0500 &0.1997\\   		
			SNNDPC&$k/NC$&17/2&-0.0030   &0.0063   &23/12&0.0042    &0.1165 \\  
			FKNN-DPC&$k/NC$&15/2&0.0000 &0.0000   &20/12&0.0000 &0.0000 \\ 
			VDPC&$pct/\delta_t$&2/20&{\bfseries 0.1399}&{\bfseries 0.1013}&1/30&{\bfseries 0.1126 }&{\bfseries 0.3657 }\\ 
			\bottomrule 
		\end{tabular}  
The best results are highlighted in boldface.	
	\end{threeparttable}
\end{table}

\subsection{Experiments on Images Datasets}
Image clustering is a challenging machine learning task, which can effectively assess the performance of the clustering algorithms. We select the following six popular images datasets to show the performances of all the clustering results:

(\rmnum{1}) $COIL-20$. It is a popular greyscale image dataset from Columbia University Image Library, which comprises 1440 images. Each image has 128x128 pixels, which are viewed as 16,384 features. We take the matrix of size 1440x16384 as input data.

(\rmnum{2}) Olivetti Faces and Olivetti Faces-100.
Olivetti Faces dataset comes from AT\&T Labs, which comprises the facial images of 40 different persons (each has 10 images). Olivetti Faces-100 refers to the first 100 facial images of Olivetti Faces. Each image has 64x64 pixels, which are viewed as 4096 features. We use the image similarity measurement method CW-SSIM \cite{sampat2009complex} to compute new representations of the original images in the same way as carried out in \cite{rodriguez2014clustering}. Finally, we take the similarity matrix of 400x400 as input data for dataset Olivetti Faces and take the similarity matrix of size 100x100 as input data for dataset Olivetti Faces-100.

(\rmnum{3}) Umist Faces. It is a popular greyscale facial image dataset, which comprises 575 images collected from 20 persons. Each image has 112x92 pixels, which are viewed as 10304 features. We take the similarity matrix of size 575x10304 as input data. 

(\rmnum{4}) mini-Corel5K. It comprises 100 images, which were selected from the popular semantic RGB image dataset Corel5k. Corel5k has 50 semantic topics, where each topic has 100 images. The dataset mini-Corel5K has 10 semantic topics, where each topic has 10 images. Each image has 32x32 pixels, which are viewed as 1024 features.
we take the similarity matrix of size 100x100 as input.

(\rmnum{5}) mini-Cafar10. Cifar-10 is a small RGB image dataset to identify universal objects, which have 10 classes. The Cifar-10 dataset has 50000 training images and 10000 testing images, we select 1000 images (100 images from each class) from the testing images to form the mini-Cafar10  dataset. 

\begin{figure}[H]
	\centering
	\subfigure[  ]{
		\label{Fig1rdg}
		\includegraphics[width=0.4\textwidth]{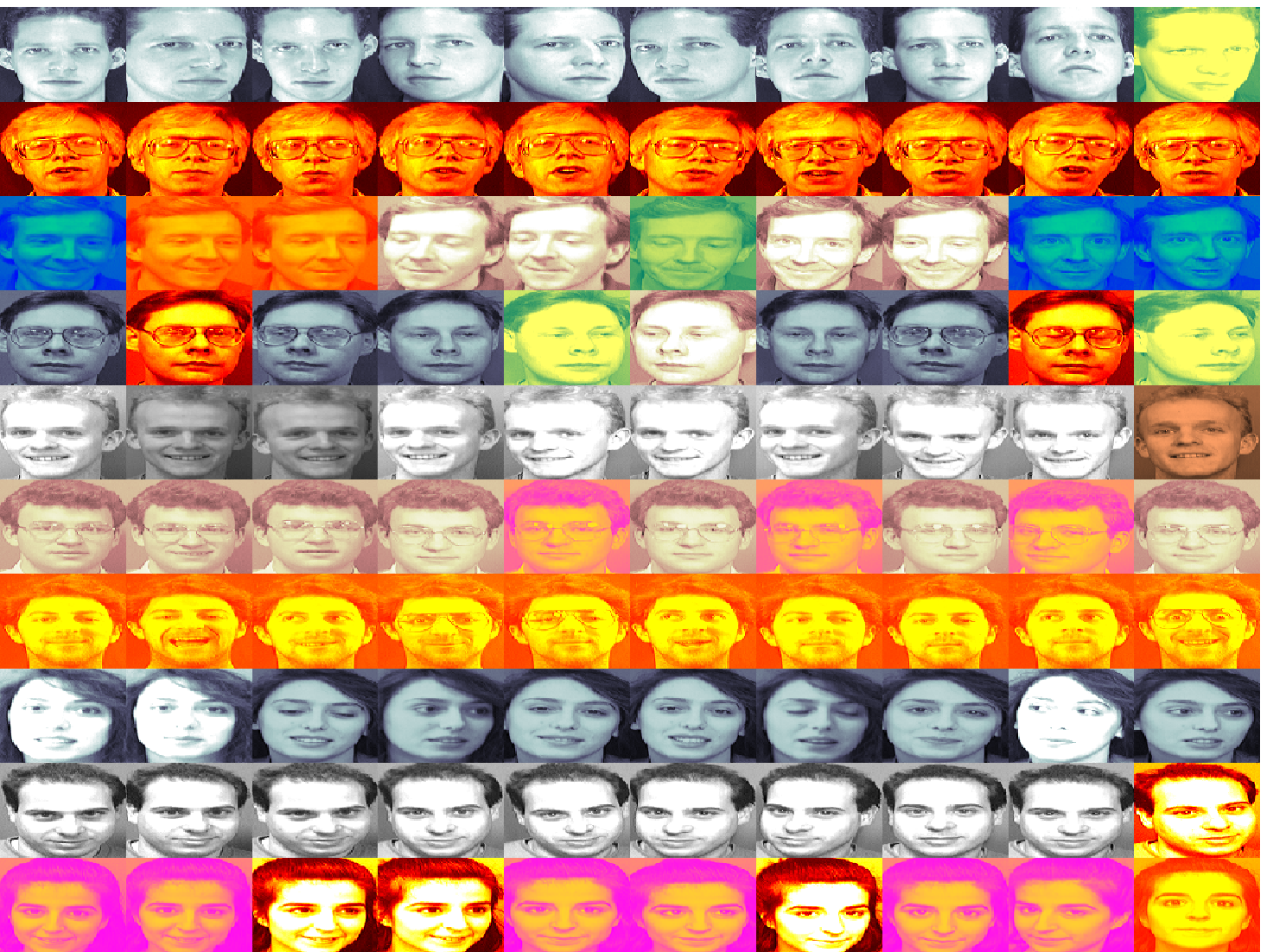}}
	\subfigure[]{
		\label{figoliver}
		\includegraphics[width=0.4\textwidth]{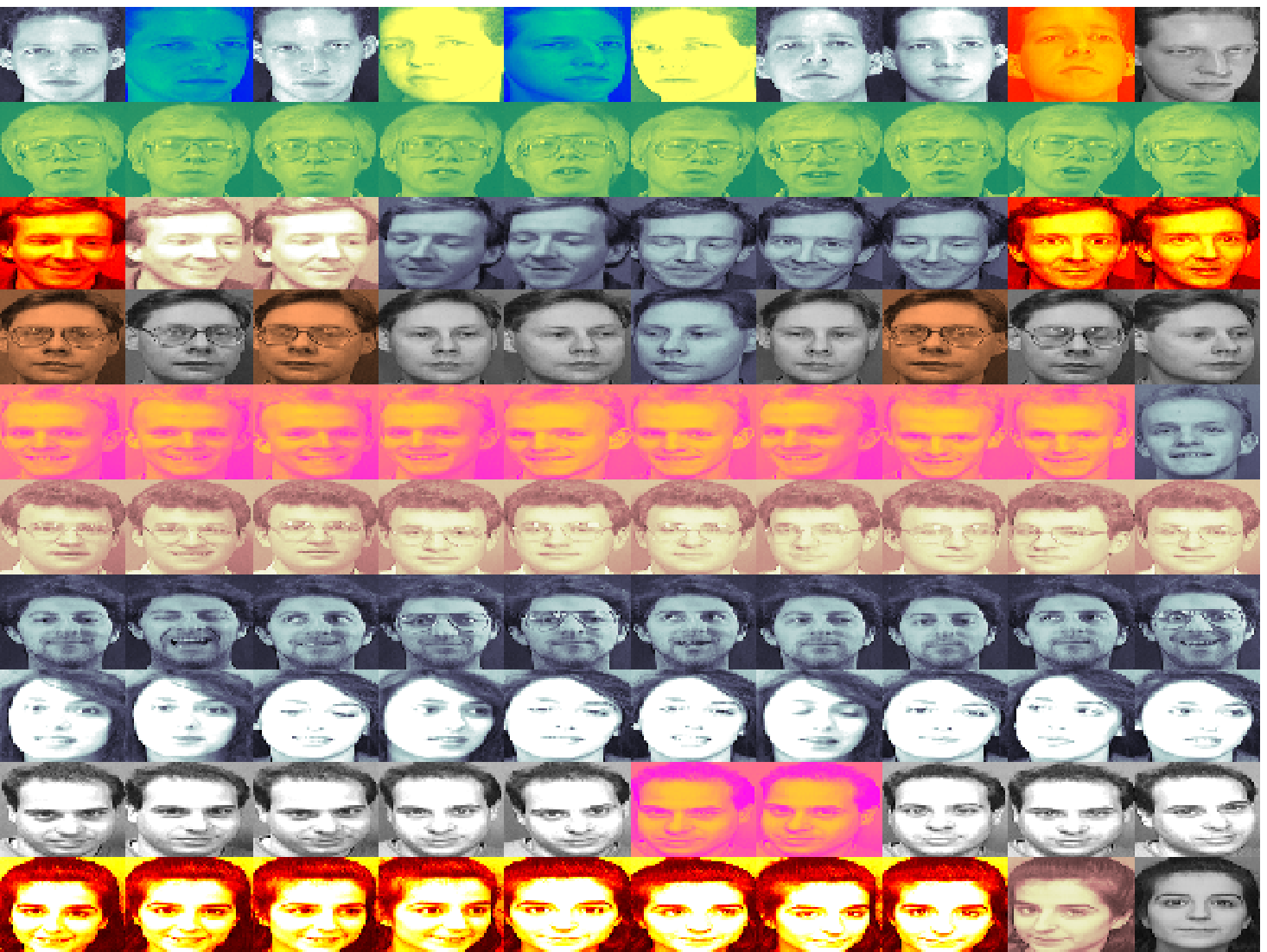}}	
	\caption{ Clustering results on dataset Olivetti Faces-100. (a) Olivetti Faces-100 clustering results generated by DPC. (b) Olivetti Faces-100 clustering results generated by VDPC.}
	\label{Figoliverall}
\end{figure}

The first four image datasets have single background and the last two image datasets have complex background. All the clustering algorihtms take the same input matrix. As shown in Table \ref{tableimage}, VDPC achieves the best clustering results on all the six image datasets. The clustering results on the Olivetti Faces-100 dataset are shown in Figure~\ref{figoliver}  for performance illustration. 

\begin{table}[htbp]
	\centering	
	\begin{threeparttable}
		\centering   
		\caption{ Comparison of six clustering algorithms on images datasets}  
		\label{tableimage}  
		\small 
		\begin{tabular}{ccccc|cccc}  
			\toprule
			Algorithms &Par&Val&ARI&NMI&Val& ARI& NMI\\
			\midrule  
			\multicolumn{5}{c|}{\bf Dataset COIL20}&\multicolumn{3}{c}{\bf Dataset Olivetti Faces-100} \\					
			DBSCAN&Eps/MinPts&9000/2&0.0000&0.0028&0.85/2&0.5918& 0.7979\\ 	
			DPC&$pct$&1&0.4809&0.7633&3&0.6023&0.7802\\
			DPC-KNN&$pct/d/k$&1/3/5&0.1862 &0.4902&1/20/5&0.6790&0.8263\\ 
			McDPC&$\gamma/\theta/\lambda/pct$&1/10/7800/1&0.2090  &0.5345 &1/0.1/0.83/100&0.5525  &0.7788  \\
			SNNDPC&$k/NC$&14/20& 0.6232  &0.8546  &5/10&0.6533 &0.7983\\ 
			FKNN-DPC&$k/NC$&12/20&0.2890 &0.4780  &6/10&0.3408     &0.6621   \\ 
			VDPC&$pct/\delta_t$&1/6300&{\bfseries 0.6306}&{\bfseries 0.8549 }&1/0.827&{\bfseries 0.7475 }&{\bfseries 0.8596 }\\ 
			\hline 		
			\multicolumn{5}{c|}{\bf  Dataset Olivetti Faces}	&\multicolumn{3}{c}{\bf Dataset UMist Faces}\\ 					
			DBSCAN&Eps/MinPts&2600/3&0.0000&0.0000&30/6&0.0000  &0.0000  \\ 		
			DPC&$pct$&1&0.4257&0.8156&100&0.3685& 0.6644 \\
			DPC-KNN&$pct/d/k$&10/3/5&0.0833&0.7412&2/6/6&0.3321 &0.6848  \\
			McDPC&$\gamma/\theta/\lambda/pct$&2/0.1/2936/100&0.4116  &0.7016   &2/1/1.6/0.2&0.1723 &0.1507\\  
			SNNDPC&$k/NC$&5/40& 0.3136   &0.7622   &8/20&0.3600    &0.6649 \\ 
			FKNN-DPC&$k/NC$&4/40&0.2677    &0.6742   &9/20&0.2722    &0.5072 \\  
			VDPC&$pct/\delta_t$&10/1583&{\bfseries  0.4702}&{\bfseries  0.8496 }&100/2936&{\bfseries0.4525}&{\bfseries 0.7296  } \\ 
			\hline
			\multicolumn{5}{c|}{\bf Dataset mini-Corel5K}&\multicolumn{3}{c}{\bf Dataset mini-Cafar10}\\ 
			DBSCAN&Eps/MinPts&3/2&0.0000&0.0000&4000/2&0.0000&0.0000\\ 
			DPC&$pct$&90&0.1609 &0.5312    &1&0.0187 &0.4089 \\
			DPC-KNN&$pct/d/k$&90/5/5&0.1505 &0.5096 &1/2/3&0.0083  &0.2120\\
			McDPC&$\gamma/\theta/\lambda/pct$&1/0.1/0.83/10&0.1296 &0.0801 &1/0.1/1370/10&0.0193  &0.3382\\   		
			SNNDPC&$k/NC$&7/10&0.1640  &0.3664   &26/10&0.0170     &0.0527 \\  
			FKNN-DPC&$k/NC$&4/10&0.1225  &0.3664   &12/20&0.0164  &0.0368  \\ 
			VDPC&$pct/\delta_t$&90/0.83&{\bfseries 0.1754}&{\bfseries 0.5621 }&10/1370&{\bfseries 0.0194  }&{\bfseries 0.4166 }\\ 
			\bottomrule 
		\end{tabular}  
	The best results are highlighted in boldface.	
	\end{threeparttable}

\end{table}

\section{Conclusion and Future Work}
\label{cafw}

In this paper, we propose an effective variational clustering algorithm called VDPC, which has only two user-defined parameters. VDPC utilizes the advantages of two well-known methods DBSCAN and DPC to perform the clustering procedures in an autonomous way according to the proposed thought of local denisty levels. We use 20 datasets to evaluate the performance of VDPC, the experimental results show that VDPC outperforms two classical clustering algorithms (DBSCAN and DPC) and four state-of-the-art extended DPC algorithms in all cases.

In the future, for the two user-defined parameters \textit{pct} and $\delta_t$ in VDPC, we will try to find a heuristic method to determine ther values. As such, it is easier for users to apply VDPC in different fields.

\section{Acknowledgment}

This research is supported by the Innovation and Enterpreneurship Program of Jiangsu Province (Grant No. JSSCBS20211048) and Jiangsu Provincial Universities of Natural Science General Program (Grant No. 21KJB520021).

\begin{appendices}
	\section{Effect of Different \textit{k} Values in aSNNC on Clustering Results}
	\label{assnnc}
The widely adopted settings of \textit{k} are $k=\lceil \sqrt{nl} \rceil$ and $k=\lceil \ln{(nl)} \rceil$.
For typical variational density datasets Compound and Pathbased, VDPC obtains the best results when $k=\lceil \sqrt{nl} \rceil$ as shown in Table \ref{ablationk}. Thus, we set $k$ in aSNNC to a heuristic value $\lceil \sqrt{nl} \rceil$.
	
	\begin{table}[htbp]
		\centering   
		\caption{  Clustering results (ARI)  of applying different $k$ values in aSNNC}  
		\label{tablenum2}  
		\setlength{\tabcolsep}{0.5mm}{
			\begin{tabular}{ccccccccccc}  
				\toprule  
				Datasets& Compound&Pathbased\\	
				\midrule  
				$k=\lceil \sqrt{nl} \rceil$&{\bfseries1.0000}&{\bfseries1.0000}\\
				$k=\lceil \ln{(nl)} \rceil$&0.7964 &{\bfseries1.0000}\\
				\bottomrule  
				\label{ablationk} 
		\end{tabular} }	
	\end{table}

	\section{Effect of Different \textit{Eps} Values in aDBSCAN on Clustering Results}
	\label{adbscan}

For two representative variational density datasets Compound and Pathbased, when we set \textit{Eps} to $Eps=sim(\lceil\sqrt{\vert C_{x_{low}} \vert}\rceil)$, VDPC obains the best clustering results as shown in Table \ref{tablenum2}. Thus, we set \textit{Eps} in aDBSCAN to a heuristic value $sim(\lceil\sqrt{\vert C_{x_{low}} \vert}\rceil)$.
	\begin{table}[htbp]
		\centering   
		\caption{  Clustering results (ARI)  of applying different \textit{Eps} values in aDBSCAN}  
		\label{tablenum2}   
		\setlength{\tabcolsep}{0.5mm}{
			\begin{tabular}{ccccccccccc}  
				\toprule  
				Datasets& Compound&Pathbased\\	
				\midrule  
				$Eps=sim(\lceil\sqrt{\vert C_{x_{low}} \vert}\rceil) $&{\bfseries1.0000}&{\bfseries1.0000}\\
				$Eps=sim(\lceil \ln({\vert C_{x_{low}} \vert})  \rceil )$&\textit{NaN}&{\bfseries1.0000}\\
				\bottomrule  
				\label{ablationeps} 
				\textit{NaN}: not number.
		\end{tabular} }
		
	\end{table}

	\section{Applying aDBSCAN and aSNNC in Different Density Levels }
	\label{dbandsnnc}
To further evaluate our intuition that aSNNC works well for lower density level $l_1$ and aDBSCAN works well for $l_p$, we conduct more experiments for all the algorithm-density level combinations and show the results in Table \ref{ablation}. For the two representative variational density datasets Compound and Pathbased, the proposed combination of aSNNC for $l_1$ and aDBSCAN for $l_p$ obtains the best results.
	
	\begin{table}[htbp]
		
		\centering   
		\caption{  Clustering results (ARI)  of applying aDBSCAN and aSNNC in different density levels}  
		\label{tablenum3}  
		\setlength{\tabcolsep}{0.5mm}{
			\begin{tabular}{ccccccccccc}  
				\toprule  
				Datasets& Compound&Pathbased\\	
				\midrule  
				aSNNC ($l_1$)+aSNNC ($l_p$) &0.5208&{\bfseries1.0000}\\
				aDBSCAN ($l_1$) +aDBSCAN ($l_p$) &0.1914 &0.0058\\
				aDBSCAN ($l_1$) +aSNNC ($l_p$) &0.4415 &0.0058\\
			aSNNC ($l_1$) +aDBSCAN ($l_p$)	({\bfseries Used in VDPC})&{\bfseries1.0000}&{\bfseries1.0000}\\
				\bottomrule  
				\label{ablation} 
		\end{tabular} }
		
	\end{table}
		
\end{appendices}

\bibliography{mybibfile}
\end{document}